# Challenges in Expanding Portuguese Resources: A View from Open Information Extraction

**Marlo Souza**  [ Federal University of Bahia | *msouza1@ufba.br* ]
**Bruno Cabral**  [ Federal University of Bahia | *bruno.cabral@ufba.br* ]
**Daniela B. Claro**  [ Federal University of Bahia | *dclaro@ufba.br* ]
**Lais Salvador** [ Federal University of Bahia | *laisns@ufba.br* ]

*Institute of Computing, Universidade Federal Fluminense, Av. Gal. Milton Tavares de Souza, s/n, São Domingos, Niterói, RJ, 24210-590, Brazil.*

**Abstract.** Open Information Extraction (Open IE) is the task of extracting structured information from textual documents, independent of domain. While traditional Open IE methods were based on unsupervised approaches, recently, with the emergence of robust annotated datasets, new data-based approaches have been developed to achieve better results. These innovations, however, have focused mainly on the English language due to a lack of datasets and the difficulty of constructing such resources for other languages. In this work, we present a high-quality manually annotated corpus for Open Information Extraction in the Portuguese language, based on a rigorous methodology grounded in established semantic theories. We discuss the challenges encountered in the annotation process, propose a set of structural and contextual annotation rules, and validate our corpus by evaluating the performance of state-of-the-art Open IE systems. Our resource addresses the lack of datasets for Open IE in Portuguese and can support the development and evaluation of new methods and systems in this area.

## 1 Introduction

Open Information Extraction (Open IE) is the task of extracting structured semantic information from unstructured sources such as documents or web pages Etzioni *et al*. [2008]. Multiple Open IE methods have been proposed in the literature for different languages, but most work in the area relies on unsupervised approaches Glauber and Claro [2018]. Recently, however, boosted by advances in machine learning and the availability of new corpora for the problem, Open IE methods based on supervised learning have been proposed Stanovsky *et al*. [2018]; Cui *et al*. [2018]; Sun *et al*. [2018]; Zhang *et al*. [2017], achieving state-of-the-art results for the English language.

As noted by Glauber and Claro [2018], major advances in Open IE have mainly focused on the English language. As pointed out by Bender Bender [2009, 2019], the focus on English and its particular characteristics may introduce bias to the field, and validation of methods in diverse languages is necessary. A cross-lingual verification is hard to perform due to the lack of available corpora for Open IE in different languages. Unfortunately, manual creation of these corpora is a difficult task Glauber *et al*. [2018]; Lechelle *et al*. [2019], partly due to the vague notion of semantic relation advocated in the area Xavier *et al*. [2015]; Lechelle *et al*. [2019]; Stanovsky and Dagan [2016].

In this work, we report our experience in the creation of a manually annotated corpus for Open Information Extraction for the Portuguese language. This annotation process was carried out in an iterative process by five human annotators with experience in linguistic annotation initiatives. Based on our methodology and a theoretically sound concept of relation adopted in this work, we believe we obtained a high-quality linguistic resource that can help the development of methods and systems for the area. Further, by using a multilingual parallel corpus as a base, namely Parallel Universal Dependencies (PUD) Nivre *et al*. [2020], our corpus can be extended to other languages and become a standard for cross-lingual evaluation of Open IE systems. The resulting OIEC-PT corpus is (anonymously) available at https://hyperalgesic-combs.000webhostapp.com/corpora.zip.

## 2 Semantic Relations and Structured Propositions

Open Information Extraction relies on the identification of semantic relations between entities expressed in natural language. While traditional Information Extraction methods rely on a pre-existing set of well-defined semantic relations relevant to a specific domain, the notion of *relation* in the Open IE literature is often vague and can vary among authors. To establish proper methodologies for evaluating methods and creating datasets, it is crucial to clarify the concepts involved. In this section, we aim to answer two questions:

i. What is the nature of semantic relations and relationships discussed in this work?
ii. How can we recognize an instance of such a relation occurring in text?

The concepts of *relation* and *relationship* are fundamental notions studied in areas such as Computer Science, Linguistics, and Philosophy. Peter Chen Chen [1976] defines a *relationship*, in the context of Entity-Relationship modeling, as an association among entities. He does not directly discuss the nature or structure of relationships in that work. For Guarino and Guizzardi Guarino and Guizzardi [2015], on the other hand, *relationships* are entities that act as truth-makers of some proposition relating two or more entities, i.e., a relation holding among *these* entities. A truth-maker is an element whose existence makes a particular proposition true.

We distinguish between *relations* and *relationships* as follows: a *relation* is an abstract, second-order type that categorizes the kinds of associations that can hold between entities, properties, or classes of entities or properties. A *relationship* is a concrete instance of a relation that holds between specific entities or properties in the world. In other words, relationships are instances of relations.

For example, consider the relation *is father of*. This is a type of association that can hold between two people. An instance of this relation is the relationship between *João* and *Maria* in the statement "João é pai de Maria"[1]. In this case, "João é pai de Maria" expresses a specific relationship, which is an instance of the more general relation *is father of*.

In our work, we focus on identifying and extracting such relationships from text, which are the actual instances of relations occurring in sentences. Recognizing relationships in text requires connecting the superficial structure of a sentence to the propositions it expresses. We adopt the notion of *structured propositions* King [1995]; Soames [2019]; Pickel [2019], which posits that propositions are structured entities and that the structure of the proposition expressed by a sentence is a function of the structure of that sentence.

By adopting structured propositions as the foundation of our work, we can argue for the soundness of syntactic-based rules for identifying relationships in text and construct such rules in a principled manner. As such, we can explain why, from a sentence such as "Laura mudou-se para a casa do lago em 1948"[2], with the dependency tree depicted in Figure 1, we can extract facts such as *Mudou_para*(*Laura*, *a casa do lago*), *Mudou_em*(*Laura*, *1948*), *Mudou_para_em*(*Laura*, *a casa do lago*, *1948*), and also the implicitly stated relationship *Localiza_em*(*a casa*, *o lago*).

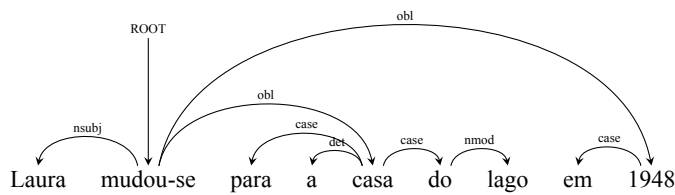

**Figure 1.** Dependency tree of the sentence "Laura mudou-se para a casa do lago em 1948"

Therefore, we define that a relationship can be recognized in a textual fragment if (i) it describes a proposition subsumed by the textual fragment, and (ii) its structure is embedded, or can be obtained by some defined sense-preserving transformation, from the structure of the sentence. With that, we can evaluate the validity of annotation rules and guides for Open IE annotation initiatives and propose a set of rules that will compose our annotation guideline.

## 3 Open IE Corpora

As noted by authors Glauber and Claro [2018] and Xavier *et al*. [2015], the area of Open IE still lacks evaluation standards, both in terms of well-accepted and widely adopted evaluation metrics and standard evaluation resources such as corpora for the task.

To our knowledge, the first work that proposed a systemic evaluation of available systems was Del Corro and Gemulla [2013]. In that work, the authors benchmarked different Open IE systems for the English language. For that, the authors semiautomatically created a dataset for the evaluation with two human judges evaluating the extractions performed by several systems. The annotation process consisted of a single step of annotation with no discussion of disagreements or further validation. As a result of the adopted annotation methodology, we believe the dataset may contain severe noise.

More recently, authors in Stanovsky and Dagan [2016] propose the OIE2016 benchmark, an extensive dataset for Open IE created from a corpus of Semantic Role Labelling He *et al*. [2015] for the English language. The validation of the resulting extractions was performed by a single human annotator on a random sample of 100 sentences. While authors point out the vague notion of relations in previous work, they fail to present a positive characterization of the concept adopted in their work and provide only a superficial guideline for the annotation process. Furthermore, as the quality estimation of the resulting dataset was performed based on only a random sample of sentences validated by a single annotator, it is not clear how to assess the quality of their results.

Based on a critique of the overall quality of the OIE2016 benchmark, authors in Lechelle *et al*. [2019] propose the WiRe57 dataset for the English language. WiRe57 is composed of 57 sentences from which 343 relations were manually extracted by two human annotators. While the authors present a more substantial discussion of the principles and guidelines adopted in the annotation process, their work relies on a vaguely defined notion of relation

Continuing with the English language, authors in Bhardwaj *et al*. [2019] propose the CaRB dataset based on a crowdsourcing strategy. The authors employed Amazon Mechanical Turk (MTurk) to annotate extractions for 1,282 sentences with workers trained by a tutorial designed to evaluate their performance in the Open IE annotation task. Although the authors employ quality control of the annotators through the designed tutorial, as Open IE is a notably difficult task, the quality of the resulting corpus is questionable, in our opinion. Authors fail to discuss specific annotation guidelines and conceptual foundations of the phenomenon being annotated. To assess the quality of the dataset, they employ three strategies based on human validation, the performance of systems trained with the corpus, and a combination of the ap-proaches.

---

[1]"John is the father of Mary"
[2]"Laura moved to the lake house in 1948"

Although Open IE has undoubtedly gained importance in the area in the last decade mainly due to its potential to use in various downstream applications, most systems and methods are still focused on the English language. An annotation corpus is crucial for fostering the development of new methods and the evaluation of the existing ones in low resources languages, such as Portuguese.

The most recent initiative to the Portuguese language, as far as our knowledge, was conducted by Glauber *et al.* [2018]. The authors present a Portuguese Open IE corpus based on a clearly defined annotation methodology and specific guidelines for the Portuguese language. Although their work is an important first step, their corpus is composed of a very limited set of sentences, only 25 from which around 400 relations were extracted. The corpus was annotated by 5 human annotators in a two-step process and evaluated based on inter-annotation agreement and qualitative discussion of the results. The authors, however, fail to present a consistent formal definition of relation as understood in their work and report that this was a significant reason for disagreement in the annotation.

Different from these works, our annotation process is rooted in a rigorous methodology and supported by a well-established semantic theory and resources with connections to well-known philosophical and linguistic traditions.

## 4 Corpus Generation

### 4.1 Methodology

One hundred sentences from the Parallel Universal Dependencies (PUD) corpus Nivre *et al.* [2020] were selected, equally distributed and manually annotated by a team comprising a senior linguist and three junior linguists, in accordance with basic general rules. The annotations made by each linguist were subsequently assessed by the others as valid or invalid. After this first evaluation, the annotators proposed other possible extractions whenever they identified them, which also went through a new evaluation cycle. During the process, the main complexities of the task were discussed, leading to the proposition of a set of rules that could guide future annotation efforts.

### 4.2 Annotation Guide

The annotation guidelines are composed of foundational concepts for this annotation effort and a set of rules to guide the human annotation. It is important to note that our annotation guidelines define appropriate constraints that make this endeavor achievable. Based on our principles, we developed a set of structural and contextual rules. We formulated a total of seven structural rules and two contextual rules to guide the annotation process.

#### 4.2.1 Structural Rules

(**S1**) A valid relation is defined by the structure *{NP, VP, NP}* or *{NP, VP, AP}*, in which the relation descriptor (VP) contains the main verb and the arguments are in a dependency relationship with the VP.

(**S1.1**) If the nominal phrase describing any of the arguments of a relation is expressed by a pronoun whose referent cannot be retrieved in the sentence, the extraction is considered invalid. However, if the referent can be retrieved from the sentence, the extraction is considered valid.

(**S2**) Relation descriptors and arguments in the extracted facts must agree in number and gender.

(**S3**) In the noun phrase that constitutes the arguments of the relation, the adjuncts that accompany the nominal nucleus must be in a dependency relation and agree with it. So, the arguments are composed of the nominal nucleus and its respective determiners and modifiers (articles, numerals, adjectives, and some pronouns).

(**S4**) When the complement or adjunct of the main verb in the relation descriptor is a prepositional phrase (PP), a valid extraction must have the preposition attached to the fragment (rel). For example, given the sentence "David viaja para outro país"[3], one valid extraction could be ⟨David, viaja para, outro país⟩.

(**S4.1**) When the complement or adjunct of the main verb in the relation descriptor is a prepositional phrase (PP) and the preposition appears in a contracted form with a determiner, this contraction must be broken—the preposition must be attached to the fragment *rel* and the determiner to the fragment *arg2*. For instance, given the sentence "David deu sua bagagem ao motorista"[4], one valid extraction could be ⟨David, deu sua bagagem a, o motorista⟩.

(**S4.2**) However, if the preposition integrates a lexical expression, i.e., a sequence of words with one unit of meaning, such as "ao longo de" ("along") and "com foco em" ("focusing on"), the extraction that fragments the locution is considered invalid. So, an extraction like ⟨um trabalhador, anda a, o longo do trem⟩ is invalid, since it fragments the lexical expression and compromises the meaning.

(**S5**) When a sentence has adverbs modifying verbs or adjectives, in a valid extraction, the adverb must appear together with the verb or adjectives it modifies. For instance, given the sentence: "Os governantes assumiram publicamente a responsabilidade pelas mortes"[5], one possible extraction is ⟨Os governantes, assumiram publicamente, a responsabilidade pelas mortes⟩, where the verb and the adverb appear together in the relation descriptor.

#### 4.2.2 Contextual Rules

(**R4**) An extraction ⟨$arg_1, rel, arg_2$⟩ will be considered valid only if the nucleus of the noun phrase representing the entity (in the original sentence) referred by the argument ($arg_1$ or $arg_2$) is included in the respective argument. An extraction is said to be *informative* when the semantic relation expressed by the triple corresponds accurately to the information presented in the sentence, without introducing or omitting essential elements.

---

[3]"David travels to another country"
[4]"David gave his luggage to the driver"
[5]"Government officials publicly claimed responsibility for the deaths"

Table 1. Inter-annotator agreement in all annotation steps

|  | 1st step | 2nd step | 3rd step |
|---|---|---|---|
| Number of judges | 5 | 2 | 3 |
| Randolph's Kappa | 0.63 | 0.86 | 0.94 |
| Percentage of agreement | 0.81 | 0.93 | 0.97 |

For example, given the sentence "O símbolo de estrela do PT vai emoldurar o cenário dos programas do candidato Luiz Inácio Lula da Silva"[6], the extraction ⟨PT, vai emoldurar, Luiz Inácio Lula da Silva⟩ is well-formed structurally but uninformative because "PT" and "Luiz Inácio Lula da Silva" are not the nuclei of their respective NPs in the original sentence, and the extraction does not accurately reflect the relationship implied by the sentence.

### 4.3 Corpus Characteristics

The Open IE Corpus - PT (OIEC-PT) is composed of the first 300 sentences of the Portuguese section of the Parallel Universal Dependencies (PUD) corpus Nivre *et al*. [2020]. The PUD corpus is a written-language corpus composed of 1,000 sentences extracted from news sources and Wikipedia and annotated with morphological and syntactic information following the Universal Dependencies v2 guidelines.

During the annotation process, we identified a number of sentences in the PUD treebank that contained errors in the morphosyntactic annotation. To ensure the quality and fairness of our corpus as an evaluation resource, we decided to remove these sentences from the OIEC-PT corpus. This resulted in the corpus containing only sentences with correct morphosyntactic annotations, thereby improving the reliability of the extractions and supporting more accurate evaluation of Open IE systems. We believe that this approach strengthens the validity of our work and makes the corpus more useful to the community.

### 4.4 Results

Our team of human annotators was composed of five individuals with previous experience both on the task of Open IE and on annotation of linguistic data. The team consisted of three male and two female annotators, all of them college-educated native speakers of Brazilian Portuguese of mixed ethnicity, researchers, and practitioners with experience in the field of Natural Language Processing and Computational Linguistics.

To evaluate the agreement between the human annotators, we used free-marginal multi-rater kappa Randolph [2005]. The agreements (Randolph's kappa) and percentage of agreed judgments among the annotators in each step of the annotation process are presented in Table 1.

The resulting corpus consists of two sets of sentences annotated with syntactic dependency information, obtained from the PUD treebank, and identified the relationships. The first set, called the *silver set*, is composed of 200 sentences used

---

[6]"The star symbol of PT will frame the scenario of the candidate's programs Luiz Inácio Lula da Silva"

in the first and second steps of the annotation aiming to calibrate the annotation guideline. The second set, called the *gold set*, is composed of 100 sentences annotated in the third step and is the result of the annotation process. From the set of 300 input sentences, 473 extractions were obtained after all annotation rounds, 136 of which are in the gold corpus as presented in Table 2.

Table 2. Corpus Statistics

| Characteristic | Corpus | |
|---|---|---|
|  | Silver | Gold |
| # of Sentences | 200 | 100 |
| # of Extractions | 337 | 136 |
| Average # extractions per sentence | 1.69 | 1.36 |
| Average size of extraction (number of words) | 9.83 | 8.44 |
| Average size of arg1 (number of words) | 2.98 | 2.51 |
| Average size of arg2 (number of words) | 4.56 | 3.76 |
| Average size of relation descriptor (number of words) | 2.30 | 2.17 |

## 5 Validation

Recent work, such as Cabral *et al*. [2020a,b], has proposed the use of classifiers to assess the quality of Open IE extractions to assist the creation of an Open IE dataset. We adopted them to probe the usefulness of these models to such a task, and to investigate the quality and coherence of annotations - comparing the silver and golden sets of our corpus.

### 5.1 Experimental Methodology

Aiming to evaluate the intrinsic difficulty of employing these classification methods to identify valid semantic relationships in text and annotation coherence between the silver and golden sets of our corpus, we performed three distinct evaluations. In these experiments, we used the top performing models in the literature for the Portuguese language: namely the Convolutional Neural Network model[7] of Cabral *et al*. [2020a], named here CrossOIE-CNN, and the CatboostProkhorenkova *et al*. [2018] classifier over the so-called language-independent representation from Cabral *et al*. [2020b], named here TABOIE-Catboost. Different from Cabral *et al*. [2020a], we employed the monolingual Portuguese model of BERT Devlin *et al*. [2019], as language-specific models achieve better results in various downstream tasks Nozza *et al*. [2020]; Souza *et al*. [2019]. As negative examples, we considered all extractions classified as invalid by the human judges in the annotation process described in Section 4.

In the first evaluation scenario, we performed a 10-fold cross-validation evaluation of the classifiers on the *golden set*, aiming to evaluate how difficult was to identify valid semantic relationships based on our *golden set* (PUD100). In the second evaluation scenario, we performed a 10-fold cross-validation evaluation of the classifiers on the entire corpus, consisting of the union of both the *silver* and *golden set* (PUD300). With this, we wish to evaluate whether the noise in the *silver set*, resulting from the unfinished guidelines, has a significant impact on learning to identify valid semantic relationships based on our *golden set*.

---

[7]We employed in our experiments the same hyper-parameters as described by the original authors, i.e. the model was trained on 80 epochs, with mini-batches of 4 and the CNN classifier has a convolution layer of 128 units with a kernel size of 3, followed by a Max pool layer and Dense Layer with 64 units, with a Softmax activation layer.

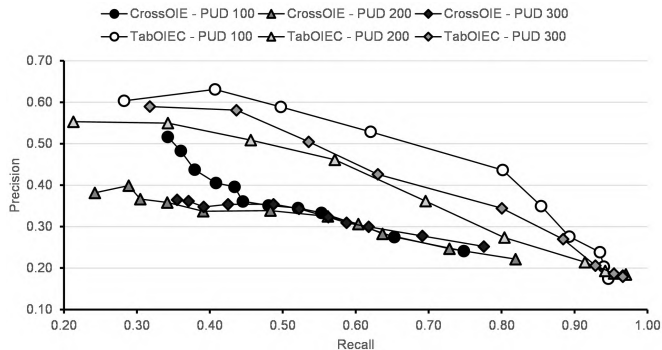

**Figure 2.** The precision-recall curves of the CrossOIE and TabOIEC classifiers for each dataset

In the third evaluation scenario, we compare the consistency of the silver set with the golden set. For that, we train the classifier in our silver set and test on the golden set (PUD200). As evaluation metrics, we compare the Area under the Curve (AUC) of the Precision-Recall curve of the classifiers, as well as Precision, Recall, F1, and Matthews correlation coefficient Matthews [1975] for specific configurations.

## 5.2 Results

To compute the Precision-Recall curve of each classifier, we employed the confidence score provided by the classifiers as a classification threshold of the method. As such, if the classifier provides a confidence score of $c$ of a relation being valid, and this value is greater or equal then the adopted threshold $t$, we consider the relation as valid. The resulting curve for the experiments PUD100, PUD200 and PUD300 is presented in Figure 2.

In Table 3, we summarize these results presenting AUC of the classifiers and well as Precision, Recall, F1, MCC and AUC for the classifier with threshold of 0.5.

According to precision and recall curves, the TabOIEC Classifier gives the best results. The AUC and MCC scores also proved this, where PUD100 on CrossOIE registered MCC of 0.307, and in TabOIEC it registered a score of 0.937.

The results presented by TabOIEC validate our progress and the methodology used in the corpus construction. PUD200 was the first version of the dataset and it have the worst AUC of 0.78. The PUD100, which was the final dataset with the consensus and the new rules based on the divergences, got the largest AUC of 0.84. Finally, the PUD300, which would be a combination of the two data sets, has a score between them of 0.81. This is an indicator that the final dataset has a higher quality than the first version, as the classifier was able to learn more.

While our experiments suggest that the higher annotation consistency in the gold set resulted in better performance of the classifiers, we acknowledge that differences in the actual texts between the silver and gold sets could also contribute to the differences in results. To fully ascertain the impact of annotation consistency versus inherent difficulty of the texts, it would be necessary to reannotate the silver set using the finalized guidelines and then re-evaluate the classifiers trained on this updated dataset. Due to time constraints, we were unable to perform this reannotation as part of the current work. We recognize this as a limitation and suggest it as an important avenue for future work.

## 6 Discussion

A challenge faced in the corpus's construction was that sentences in Portuguese could be very long. In these cases, the sentences commonly present $n$-ary relations among entities, which cannot always be reified into binary relations. The reification of $n$-ary relations was a source of errors in the annotation process that required extensive discussion among the human annotators.

In Open IE literature, while many systems focus on extracting binary relations, some systems do handle $n$-ary relations Glauber and Claro [2018]. In our work, we focused on binary relations to simplify the annotation process and because most extraction systems used in the automatic annotation step are designed for binary relations. We acknowledge that this is a limitation and that handling $n$-ary relations is an important area for future work.

Another source of disagreements in the annotation process was the detection of errors in morphosyntactic annotation in the PUD treebank. As fixing these annotations lies beyond the scope of our work, we decided to remove the affected sentences from our corpus to improve its quality and fairness.

Regarding the extraction of sentences involving pronouns whose referents can be retrieved from the sentence, we revised our rules to consider such extractions as valid. For example, in the sentence "Eu adorei as cores tropicais, diz ele"[8], the pronoun "Eu" ("I") refers to "ele" ("he") in the same sentence. Therefore, the extraction ⟨Eu, adorei, as cores tropicais⟩ is considered valid since the referent is retrievable from the sentence.

## 7 Conclusion and Future Work

In this work, we discussed the process and results of creating a high-quality, theoretically well-founded corpus for Open Information Extraction for the Portuguese language, based on the multilingual parallel treebank Parallel Universal Dependencies de Marneffe *et al.* [2014]. Our experiments showed that the iterative annotation process, aligned with a strong theoretical foundation and a well-defined set of annotation rules, ensures the quality of the process. The resulting corpus can be used to evaluate and support the creation of new systems and methods in the area for the Portuguese language.

From the observations of the annotation task, we highlight that Open IE is a difficult problem and the creation of datasets for the task should be strongly supported by both theoretical and structural constraints to guarantee a reliable result. As such, our work differs from previous attempts in the literature by being based on well-established supporting semantic theory and resources and a well-defined methodology for the annotation process. Also, as our corpus is based on a parallel treebank, we believe our annotation effort can be extended to support the creation of a multilingual dataset for (Open) Information Extraction and Entity translation, supporting the development of multilingual methods as advocated by Claro *et al*. [2019].

---

[8] "I loved the tropical colors, he says."

Table 3. Metrics scores for languages classifiers

|  |  | Precision | Recall | F1 | Accuracy | MCC | AUC |
|---|---|---|---|---|---|---|---|
| **PUD100** | CrossOIE | 0.516 | **0.342** | **0.372** | **0.819** | 0.307 | 0.66 |
|  | TabOIEC | **0.603** | 0.282 | 0.358 | 0.788 | **0.937** | **0.84** |
| **PUD200** | CrossOIE | 0.381 | **0.242** | 0.292 | 0.791 | 0.186 | 0.68 |
|  | TabOIEC | **0.553** | 0.212 | **0.293** | **0.826** | **0.947** | **0.78** |
| **PUD300** | CrossOIE | 0.343 | 0.522 | 0.403 | 0.735 | 0.255 | 0.74 |
|  | TabOIEC | **0.426** | **0.630** | **0.503** | **0.787** | **0.985** | **0.81** |

As future work, we intend to expand the corpus by annotating the remaining sentences in the PUD treebank for Portuguese, as well as extending the annotation to other languages, such as English and Spanish. Additionally, we aim to reannotate the silver set using the finalized guidelines to assess the impact of annotation consistency and further improve the resource for the community.